\documentclass[sigconf]{acmart}

\usepackage{balance}

\copyrightyear{2023}
\acmYear{2023}
\setcopyright{rightsretained}
\acmConference[MM '23]{Proceedings of the 31st ACM International Conference on Multimedia}{October 29-November 3, 2023}{Ottawa, ON, Canada}
\acmBooktitle{Proceedings of the 31st ACM International Conference on Multimedia (MM '23), October 29-November 3, 2023, Ottawa, ON, Canada}
\acmDOI{10.1145/3581783.3612708}
\acmISBN{979-8-4007-0108-5/23/10}

\makeatletter
\gdef\@copyrightpermission{
   \begin{minipage}{0.3\columnwidth}
     \href{https://creativecommons.org/licenses/by-nd/4.0/}{\includegraphics[width=0.90\textwidth]{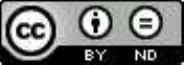}}
   \end{minipage}\hfill
   \begin{minipage}{0.7\columnwidth}
     \href{https://creativecommons.org/licenses/by-nd/4.0/}{This work is licensed under a Creative Commons Attribution-NoDerivs International 4.0 License.}
   \end{minipage}
   \vspace{5pt}
}
\makeatother

\begin{document}

\title[Jurassic World Remake]{Jurassic World Remake: Bringing Ancient Fossils Back to Life via Zero-Shot Long Image-to-Image Translation}


\author{Alexander Martin}
\email{amart50@u.rochester.edu}
\affiliation{%
  \institution{University of Rochester}
  \city{Rochester}
  \state{NY}
  \country{USA}
  \postcode{14627}}

\author{Haitian Zheng}
\email{hzheng15@ur.rochester.edu}
\affiliation{%
  \institution{University of Rochester}
  \city{Rochester}
  \state{NY}
  \country{USA}
  \postcode{14627}}

\author{Jie An}
\email{jan6@cs.rochester.edu}
\affiliation{%
  \institution{University of Rochester}
  \city{Rochester}
  \state{NY}
  \country{USA}
  \postcode{14627}}
  
\author{Jiebo Luo}
\email{jluo@cs.rochester.edu}
\affiliation{%
 \institution{University of Rochester}
 \city{Rochester}
  \state{NY}
  \country{USA}
  \postcode{14627}}

\renewcommand{\shortauthors}{Alexander Martin, Haitian Zheng, Jie An, \& Jiebo Luo}
\newcommand{\martin}[1]{\textcolor{blue}{[Martin: #1]}}
\newcommand{\luo}[1]{\textcolor{red}{[Luo: #1]}}
\newcommand{\zheng}[1]{\textcolor{red}{[Zheng: #1]}}
\newcommand{\an}[1]{\textcolor{red}{[An: #1]}}

\begin{abstract}

With a strong understanding of the target domain from natural language, we produce promising results in translating across large domain gaps and bringing skeletons back to life. In this work, we use text-guided latent diffusion models for zero-shot image-to-image translation (I2I) across large domain gaps (longI2I), where large amounts of new visual features and new geometry need to be generated to enter the target domain. Being able to perform translations across large domain gaps has a wide variety of real-world applications in criminology, astrology, environmental conservation, and paleontology. In this work, we introduce a new task Skull2Animal for translating between skulls and living animals. On this task, we find that unguided Generative Adversarial Networks (GANs) are not capable of translating across large domain gaps. Instead of these traditional I2I methods, we explore the use of guided diffusion and image editing models and provide a new benchmark model, Revive-2I, capable of performing zero-shot I2I via text-prompting latent diffusion models. We find that guidance is necessary for longI2I because, to bridge the large domain gap, prior knowledge about the target domain is needed. In addition, we find that prompting provides the best and most scalable information about the target domain as classifier-guided diffusion models require retraining for specific use cases and lack stronger constraints on the target domain because of the wide variety of images they are trained on.

\end{abstract}

\ccsdesc[500]{Computing methodologies~Computer vision}
\ccsdesc[300]{Computing methodologies~Natural language processing}

\keywords{image-to-image translation, large domain gap, stable diffusion}

\begin{teaserfigure}
  \centering
  \includegraphics[width=\textwidth, height=.3\textheight]{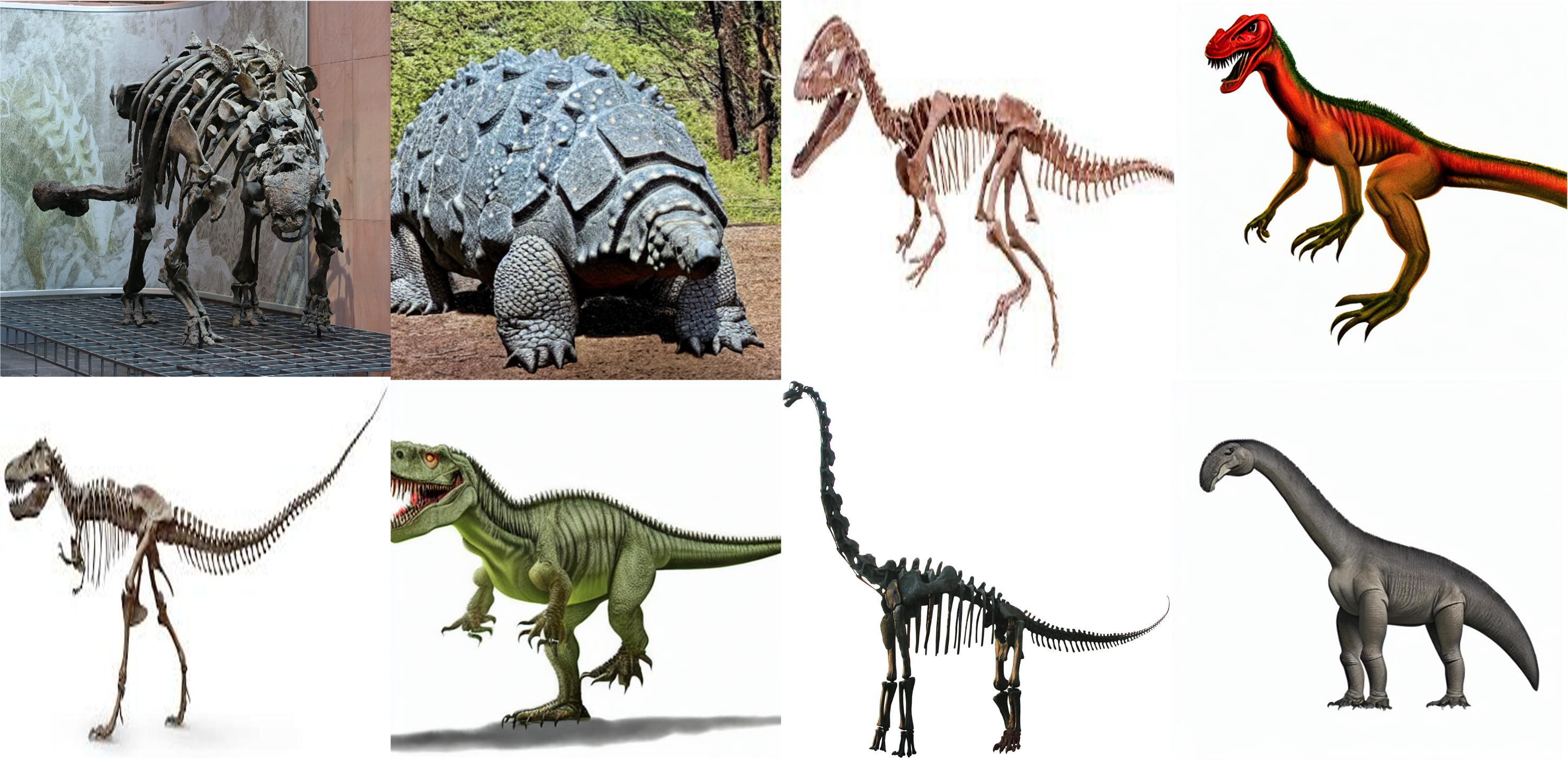}
  \caption{
  We present Revive-2I, a zero-shot image-to-image translation method that transforms given fossils images (columns 1 and 3) respectively into images of living animals of specified species based on input prompts, such as \texttt{Dinosaurs} (columns 2 and 4).
  }
  \label{fig:teaser}
\end{teaserfigure}

\received{June 07 2023}
\received[accepted]{July 16 2023}

\maketitle

\section{Introduction}
Image-to-image translation (I2I) refers to the task of translating an input image from a given source domain into a given target domain. Translating between domains is an important problem with a wide range of applications in computer vision or computer graphics like image inpainting \cite{9156579, saharia2021image}, medical imaging and disease diagnosis \cite{chen2022unsupervised, kong2021breaking, zhang2019harmonic}, image editing \cite{hertz2022prompttoprompt, mokady2022nulltext, parmar2023zeroshot, press2020emerging}, and style transfer \cite{https://doi.org/10.2312/cgvc.20221165, gatys2015neural, lin2022interactive}.

In literature, I2I has been explored under a variety of settings, including paired \cite{isola2018imagetoimage, wang2018highresolution, zhou2021cocosnet}, unpaired \cite{choi2018stargan, gokaslan2019improving, sasaki2021unitddpm, su2023dual, zhao2021unpaired, zhu2020unpaired}, and few-shot \cite{lin2020tuigan, Snell2017PrototypicalNF} image translations, with generative models such as Generative Adversarial Networks (GANs)~\cite{goodfellow2014generative}, diffusion models~\cite{sohldickstein2015deep}, and text-guided diffusion models~\cite{rombach2022highresolution}.
Despite the promising results in generating high-quality samples, the existing works mainly focus on translating between domains of images with small domain gaps, i.e., translating from photos to paintings or translating different types of animals (zebras to horses, cats to dogs). While those tasks do not require generating very different new visual features or inferences about shape during the translation process, this oversimplified setting may not reflect many practical use cases, such as translating cats to humans \cite{zhao2021unpaired}, or pumpkins to volcanoes \cite{amodio2019travelgan}, that require translating images across domain gaps with large geometry or semantic shift, i.e., \textbf{\emph{long I2I}}. 

Some longI2I tasks, like translating between animals and humans, lack verifiable constraints on the translation process. Because these translation processes are fictitious, with no potential ground truth, models can randomly learn any potential mapping as long as the result is similar to the target domain. To offer better constraints to the generation process, image editing tasks \cite{hertz2022prompttoprompt, mokady2022nulltext, parmar2023zeroshot} have been proposed to provide local edits to an image, like changing the subject of the image without changing the background. These tasks still require the generation of new geometry or a semantic shift, but provide a stricter constraint for acceptable translations.




Being able to perform constrained longI2I, longI2I with verifiable results that limit what is considered a valid translation, is an important feature for any I2I model. A model that is capable of longI2I could be used by law enforcement, taking a sketch of a perpetrator and providing a realistic photo of the person to help identify them \cite{Kim_2023_WACV, zhu2020unpaired}. It could be used by wildlife conservationists to show the effects of climate change on cities, ecosystems, and habitats \cite{Li_2015, Yang_Xu_Luo_2018, 10.1609/aaai.v33i01.3301825} or to show the hazards and impacts of wildfires \cite{jiang2021static} on towns and in nature. Or it could be used by paleontologists to translate the ancient fossils of dinosaurs and other extinct animals into their living counterparts. 


To ground these claims, we introduce and explore the task of translating skulls into living animals (Skull2Animal). This task requires generating a large amount of new visual features, generating new textures and colors. It also requires the models to make inferences about the geometry in the target domain, having to understand how to fill in parts of the animal with fat and muscle while some parts stay tighter to the skull. Unlike previous tasks from \cite{amodio2019travelgan, zhao2021unpaired}, the translation process is {\bf not fictitious}, providing {\bf verifiable} results which lead to a more constrained generation process and building off the tasks of \cite{mokady2022nulltext, parmar2023zeroshot} the translation process requires extensive inference about geometry. 

    
Unguided GAN-based methods like CycleGAN \cite{zhu2020unpaired} and ACL-GAN \cite{zhao2021unpaired} are limited in longI2I tasks because they lack an understanding of the target domain. To add guidance, Dual Diffusion Implicit Bridges (DDIB) \cite{su2023dual} uses two classifier-guided diffusion models \cite{dhariwal2021diffusion} to translate images between ImageNet \cite{5206848} classes. The classifiers provide the diffusion process with knowledge about the target domain, but the classifiers are limited to the ImageNet classes they are trained on, requiring retraining for new classes and uses.


Instead of using classifier guidance, we propose replacing the classifier with prompts. This has been previously done by null-text inversion \cite{mokady2022nulltext}, where they edit an image by modifying the prompt used to create the image. By generating the new image with a natural language understanding of what should change between each image, they successfully provide local and global edits to an image with text. 

To demonstrate the power of natural language for longI2I, we propose Revive-2I for zero-shot I2I by prompting pre-trained latent diffusion models. This method is based on the translation process of DDIB but makes three changes to the methodology by \textbf{1)} performing the diffusion process in the latent space, \textbf{2)} replacing the need for a trained classifier for guidance with a text prompt, and \textbf{3)} performing partial steps of the full forward diffusion process. These changes make the diffusion process faster, able to incorporate a broader range of classes, and robust to different I2I tasks. With a strong understanding of the target domain from natural language, we produce promising results in translating across large domain gaps and bringing skeletons back to life. The dataset and code are available at \url{https://tinyurl.com/skull2animal}. 

To summarize, our contributions are three-fold: 
\begin{itemize}
\vspace{-0.05in}
    \item We propose a novel longI2I task Skull2Animal, translating skulls into living animals.
    \item We baseline existing image-to-image translation methods (CycleGAN, ACL-GAN, DDIB, Null-text Inversion) for this task. 
    \item We propose Revive-2I to perform zero-shot I2I via prompting latent diffusion models.
\end{itemize}

\section{Related Work}
\paragraph{GANs for UI2I} Unpaired image-to-image translation (UI2I) has seen many GAN models introduced centered around a cycle consistency loss \cite{liu2018unsupervised, kim2017learning, yi2018dualgan, zhu2020unpaired}. Cycle consistency constrains the generation process so that a translated image is able to be translated back into the original image. These models have been found to do well in I2I tasks, like style transfer, that require that the core content and structure be retained during translation. However, when tasked with removing objects or performing changes to the structure of an image, models like CycleGAN \cite{zhu2020unpaired} or UNIT \cite{liu2018unsupervised} have been shown to retain artifacts from the original image because of the strict pixel-level constraint \cite{hu2020unsupervised, zhao2021unpaired}. To circumvent the strict constraint of cycle consistency and its drawbacks, methods have been introduced to perform UI2I on tasks that require more complex mappings, beyond style and texture. ACL-GAN \cite{zhao2021unpaired} introduces an adversarial cycle consistency constraint. This loss removes the pixel-level constraint of CycleGAN, removing the need to be translated back into the exact image, but still retaining important source features.

\paragraph{Diffusion Methods for UI2I} Recently, pixel-based diffusion models have been shown to be used for cycle-consistent image-to-image translation. Translation with Dual Diffusion Implicit Bridges (DDIBs) \cite{su2023dual} requires two different probability flow ordinary differential equations (PF, ODE). The source ODE converts the input images into the latent space and the target ODE synthesizes the image in the target domain. They denote this conversion ODESolve as the mapping from $x(t_0)$ to $x(t_1)$:
\begin{equation}
    ODESolve(x(t_0);v_\theta,t_0,t_1) = x(t_0) + \int_{t_0}^{t_1} v_\theta(t,x(t)), dt
\end{equation}
To perform image-to-image translation, DDIB uses two separate diffusion models. Starting with an image $x^s$ in the source domain, the latent representation, $x^l$, is produced using $x^l = ODESolve(x^s;v^s_\theta,0,1)$. Then the latent representation can be used again in a different ODESolve to obtain the target image $x^t = ODE(x^l, v^t_\theta,1,0)$. In their work, they prove that DDIBs offer exact cycle consistency, an important feature in unpaired image-to-image translation.

\paragraph{Text-Based Diffusion Methods for UI2I}
While diffusion models are able to perform UI2I, adding new classes to the methods requires retraining the diffusion models or their classifier guidance. This is computationally expensive, reducing the scalability of diffusion methods. Instead of training new models, encoding natural language with models like CLIP \cite{radford2021learning} or ALIGN \cite{jia2021scaling} can be used to guide the diffusion process in tasks like text-to-image synthesis \cite{ramesh2022hierarchical, rombach2022highresolution, saharia2022photorealistic, yu2022scaling}. These models have shown the ability to generate high-quality visual features by conditioning on text prompts. 

In this work, we build off of stable diffusion \cite{rombach2022highresolution}, a text-guided latent diffusion model. Stable diffusion performs the diffusion process in a latent space, which is perceptually equivalent to pixel-based diffusion, but more computationally suitable for the diffusion process. In the latent space, they additionally model the conditional distribution between a latent and a text prompt to perform text-to-image synthesis.

\paragraph{Image Editing Methods for UI2I} Image editing with text-guided diffusion models was first proposed by Prompt-to-Prompt \cite{hertz2022prompttoprompt}. Prompt-to-Prompt performs an intuitive image editing (translation) by editing prompts that correspond to generated images to create new images with the change in the prompt reflected in the new image. Building off of Prompt-to-Prompt, null-text inversion \cite{mokady2022nulltext} uses DDIM inversion to be able to edit real images with the Prompt-to-Prompt technique. Taking a real image and an associated caption as input, the image is inverted with a DDIM model to give a diffusion trajectory. The trajectory can then be used for null-text optimization that reconstructs the input image. With this optimization done, the image can then be edited using the technique from Prompt-to-Prompt.

\section{Dataset}
\begin{figure}[t]
    \centering
    \includegraphics[width=\linewidth]{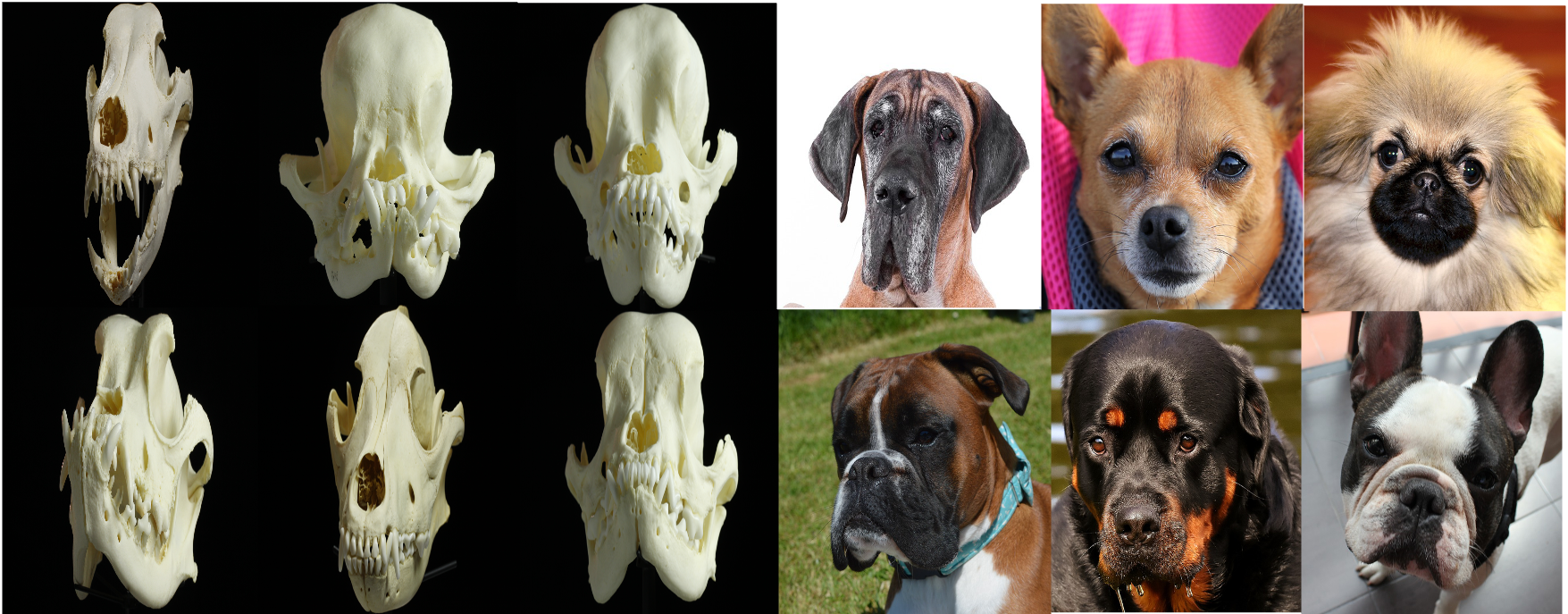}
    \caption{Example images from the collected Skull2Dog dataset.}
    \label{fig:dataset}
\end{figure}

As depicted in \autoref{fig:dataset}, our collected Skull2Animal dataset comprises unpaired images of skulls and living animals. These images were curated through a process wherein skull photographs, taken by a private photographer\footnote{Skull Photos courtesy of Nick Mann, used with permission}, were selected for inclusion. Each skull image in the dataset was captured through a comprehensive 360-degree rotation around the skull, from which a subset was manually selected by one of the authors. The selection criteria stipulated that the skull must be oriented in a manner compatible with the perspectives captured in the Animal Faces-HQ (AFHQ) dataset \cite{choi2020stargan}. This led to the aggregation of skull images aligned 90\textdegree left and 90\textdegree right from a forward-facing perspective. With the skull images collected, corresponding living animals needed to be collected. The Skull2Animal dataset consists of 4 different types of mammals: dogs (Skull2Dog), cats (Skull2Cat), leopards (Skull2Leopard), and foxes (Skull2Fox). The living animals of the dataset are sampled from the AFHQ. The dataset is partitioned by ImageNet class using a ResNet \cite{he2015deep} classifier trained on ImageNet1k \cite{5206848}.


In this paper, we will focus on the Skull2Dog subset as it provides the most skull images. In the Skull2Dog dataset, there are 6 different skulls: Boston Terrier, Boxer, Chihuahua, Great Dane, Pekingese, and Rottweiler. After an initial pass through AFHQ, there are 422 dog images that fall into those classes. To add additional images to the split, 799 dog images were randomly selected from the remainder of the AFHQ dataset, giving 1201 skull and dog images. This collection of skull and dog images was then randomly split into a training set of 1080 images and a test set of 121 test images. A more in-depth breakdown of the data partitions and breed counts can be found in \autoref{append:data}. 
\section{Methodology}
\label{sec:method}
Recently, DDIB~\cite{su2023dual} proposes leveraging a pre-trained diffusion model to define and solve ordinary differential equations (ODE) for image-to-image translation. DDIB shows promising results in transferring between predefined classes. However, as the diffusion process is conditioned on limited classes from ImageNet, DDIB cannot transfer between fine-grain and arbitrary domains, such as images of an endangered specie and its fossil. Moreover, DDIB sometimes introduces misalignment due to the lack of cross-domain constraints. Finally, diffusion on pixel space is computationally exhaustive. To address those limitations, we propose Revive-2I, a flexible and efficient framework for long I2I. Specifically,
\textbf{1)} we perform the diffusion process in the latent space with stable diffusion \cite{rombach2022highresolution}, \textbf{2)} we replace the need for a trained classifier for guidance with a text prompt, and \textbf{3)} we do not perform the full forward process. These changes make the Revive-2I translation process much quicker, able to incorporate a broader range of classes (anything that can be describe in natural language), and robust to different I2I tasks. The Revive-2I process can be broken down into two steps: encoding and text-guided decoding. 

\begin{figure} [ht]
    \centering
    \includegraphics[width=.49\textwidth, height=.15\textheight]{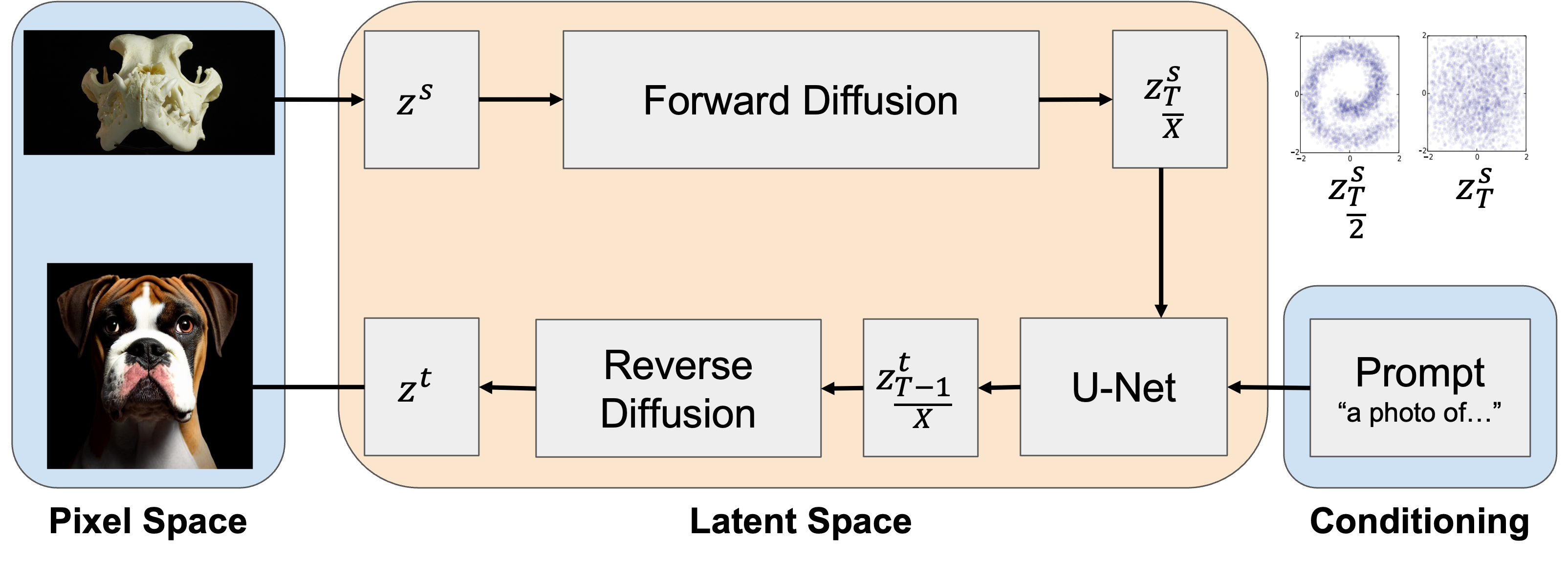} 
    \caption{Our Revive-2I for zero-shot long image-to-image translation. Our method takes a source domain (top left) image and a text prompt as input and outputs a target domain image (bottom left).
    }
    \label{fig:enter-label}
\end{figure}

\paragraph{Encoding:} Starting with a source image $x^s \in \mathbb{R}^{H \times W \times 3}$, we encode the $x^s$ into the latent representation $z^s = \mathcal{E}(x^s)$, where $z^s \in \mathbb{R}^{h \times w \times 3}$ ($h= \frac{H}{f}, w = \frac{W}{f})$ downsampled by factor $f=8$. By doing this we operate in a perceptually equivalent, but computationally more suitable space for the diffusion process \cite{rombach2022highresolution}. With the latent representation of the source image, we then apply the forward process to obtain the noised encoding $z^s_T$. This forward step is equivalent to $ODESolve(x^s;v^s_\theta,0,1)$ of DDIB, but in the latent space.

\paragraph{Varied Encoding} One might argue that any I2I task is possible with text-guided diffusion models as long as the right prompt is used. Instead of trying to find the best prompt for the task, we try taking different numbers of steps in the forward process. To convert the latent source representation, $z^s$, into the fully noised Gaussian distribution, we take $T=100$ steps of the forward process. To create partiala encodings, we take $\frac{50}{100}$, $\frac{60}{100}$, $\frac{70}{100}$, $\frac{80}{100}$, $\frac{90}{100}$, and $\frac{95}{100}$ forward steps. Taking partial steps in the forward diffusion process allows the translation process to better preserve the source content. This makes the generation more faithful to the source content but allows for the text prompt to inject the features of the target domain. Please refer to \autoref{append:shortI2I} for an applied proof of the varied encoding.

\paragraph{Prompting:} To incorporate the target domain into the diffusion process, we replace the classifier guidance from DDIB with a text prompt. If classifier guidance were used for the Skull2Animal translation process, the classifier would need to be trained for new target domains constrained to dog heads. In contrast, using text guidance, there is no training needed to perform the Skull2Animal task making the Revive-2I method able to be used across a large variety of skull images, even ones not included in ImageNet1K. The text prompt also allows for the method to constrain the generation process better. A classifier trained on ImageNet will have seen a wide variety of images under a single label. This might lead to unwanted artifacts or additional content generated in the images, like people or the full dog body. However, with a text prompt, the diffusion process can be more constrained to the ideal target domain.

The prompt used for Skull2Dog translation is ``a photo of the head of'' combined with the ImageNet label. For example, translating the Boston terrier skull would use the prompt ``a photo of the head of a Boston terrier dog.'' This prompt provides two guides to the translation process. First, we provide the same class-level guidance in DDIB by using the ImageNet class name. Second, we better constrain the generation process to only produce dog heads.

\paragraph{Guided Decoding:} We leverage the pretrained Stable Diffusion \cite{rombach2022highresolution} for text-guided decoding. Specifically, to combine the text prompt $y$ with the latent representation $z^s_T$, we first project the text prompt using the same text encoder as stable diffusion  $\tau_\theta$. This projects $y$ to an intermediate representation $\tau_\theta(y) \in \mathbb{R}^{M\times d_\tau}$. Then this intermediate representation is mapped to intermediate layers of the UNet used for denoising via a cross-attention layer. So by taking a prompt $y$ and latent representation $z^s_T$, the first step of the denoising process through UNet guides $z^s_T$ to the target domain as $z^t_{T-1}$. With the latent representation in the target domain, it can be further denoised by the reverse diffusion process giving the final latent representation $z^t$. The combination of the UNet denoising and reverse process is equivelent to $ODE(x^l, v^t_\theta,1,0)$ of DDIB in the latent space. From the latent $z^t$, we can decode the representation into the pixel space $\Tilde{x}^t = \mathcal{D}(z^t)$ resulting in an image $\Tilde{x}^t$ in the target domain.

\paragraph{Experimental Setup:} To evaluate the Skull2Dog translation process, we split the data input 1080 samples for training and 121 samples for testing. We train CycleGAN~\cite{zhu2020unpaired}, and ACL-GAN~\cite{zhao2021unpaired} on these splits to test the GAN methods. For the guided diffusion method, we use the 256x256 classifier and diffusion models from \cite{dhariwal2021diffusion} and follow the same method from DDIB~\cite{su2023dual} taking 1000 steps in the diffusion process to translate the images. For image editing we use the initial prompt ``a photo of the skull of" and the same target prompt used for Revive-2I. To evaluate the generation results, we use Fr\'echet inception distance (FID) \cite{heusel2018gans, Seitzer2020FID}, Kernel inception distance (KID) \cite{bińkowski2021demystifying} and two different top-1 classification, top-1 all and top-1 class, accuracies using ResNet-50 \cite{he2015deep} trained for ImageNet1K \cite{5206848}. Top-1 all (All @1) is scored with the top-1 classification being any of the 100 dog breeds used in ImageNet1K and top-1 class (Class @1) is scored with the top-1 classification being the correct dog breed for the specific skull. When these are scored at 100\%, top-1 all has every image classified as a dog and top-1 class has every image classified as the correct breed.

\section{Results}
\label{sec:results}
\begin{figure*}[ht]
    \centering
    \includegraphics[width=\textwidth, height=.45\textheight]{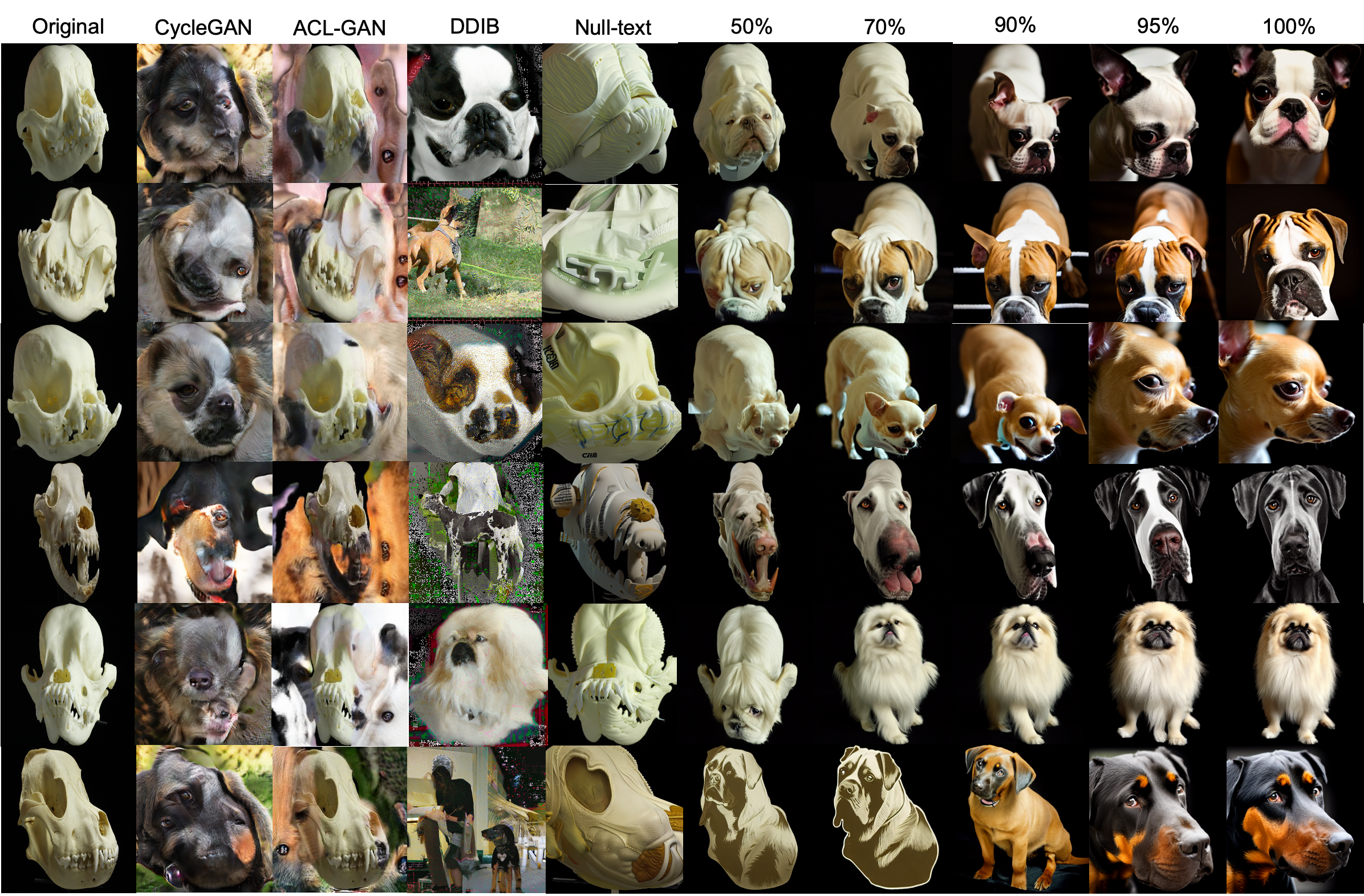}
    \caption{Skull2Dog Translations \textbf{Rows:}(top to bottom) Boston terrier, Boxer, Chihuahua, Great Dane, Pekingese, Rottweiler} 
    \label{fig:skull2dog}
\end{figure*}

\begin{table}[t]
    \centering
        \caption{Quantitative evaluation on Skull2Dog. Lower scores are better for metrics with down arrows ($\downarrow$), and vice versa.}
    \begin{tabular}{c|c|c|c|c}
         & FID $\downarrow$ & KID$\downarrow$ & All @1$\uparrow$ & Class @1$\uparrow$\\
         \hline
        CycleGAN~\cite{zhu2020unpaired} & 212.66 & 0.1325 & 77.00 & 5.70 \\
        ACL-GAN~\cite{zhao2021unpaired} & 342.16 & 0.3134 & 38.84 & 1.65 \\
        DDIB~\cite{su2023dual}  & 204.42 & 0.0698 & 73.68 & 56.14\\
        Null-text~\cite{mokady2022nulltext} & 308.19 & 0.1559 & 1.79 & 1.79 \\
        \hline
        Revive-2I$_{50\%}$ & 236.59 & 0.118 & 43.80 & 20.66 \\
        Revive-2I$_{60\%}$ & 177.30 & 0.0584 & 70.25 & 34.70 \\
        Revive-2I$_{70\%}$ & 155.066 & 0.0437 & 93.89 & 59.50 \\
        Revive-2I$_{80\%}$ & 147.85 & 0.0368 & \textbf{100} & 67.77 \\
        Revive-2I$_{90\%}$ & 149.42 & 0.0365 & \textbf{100} & 89.26 \\
        \textbf{Revive-2I$_{95\%}$} & \textbf{143.29} & \textbf{0.0345} &  \textbf{100} & \textbf{92.56} \\
        Revive-2I$_{100\%}$ & 155.9 & 0.0391 & \textbf{100} & \textbf{92.56} \\
    \end{tabular}
    \label{tab:eval}
\end{table}

The unguided GAN methods (CycleGAN and ACL-GAN) are unable to successfully perform the Skull2Animal task. As seen in \autoref{fig:skull2dog} and \autoref{tab:eval}, CycleGAN is capable of learning the gist of a dog. The model can produce features that are characteristic of dogs like the nose, eyes, and fur, but it cannot properly structure the features or understand the direction the dog is facing. In \autoref{tab:eval}, the top-1 all score for CycleGAN is high because it learns this gist of the dog and dog-like features, but its top-1 class score is incredibly low because it cannot structure the features it learns to represent a specific breed or similar looking breed depending on the skull. ACL-GAN is unable to perform the translation. The adversarial cycle consistency loss helps retain the important features from the source image in other I2I tasks, but when translating across large domain gaps, the loss retains the entire skull. This leads to the model learning to place dog like features in the background of the image and retain the skull in the center. This is able to confuse a classification model to classify some images as dogs, but because of the lack of structure, it cannot fall under the correct classes.

The image editing method, null-text inversion, is unable to create enough new visual features to translate from skulls to living animals. In some cases the editing process is able to create dog like fur textures or place a nose on the skull, but it is not able to step too far from the source image to generate new geometry. While this might be a useful feature in translation tasks between similar images, it restricts the longI2I process leading to low top-1 all and class scores. 

The guided diffusion methods (DDIB and Revive-2I) are able to translate between skulls and animals because of the understanding they have of the target domain. DDIB's classifier guidance gives the model a good understanding of the target domain allowing the model to successfully translate into the domain of living animals. However, the classifier has seen a wide variety of images under the same class label. Because of this, the translation process can result in the full body of the dog or other subjects like people in the scenes. Fixing this would require retraining the classifier for new class labels specific to the dog head use case. 

Revive-2I's text prompt allows for a {\bf more constrained translation} process without the need to retrain the diffusion model or classifier. However, this constraint is not perfect as every diffusion step from 50\% to 100\% can be shown to produce full-body dogs. We find that the model best capable of translating from skulls to living animals is Revive-2I with 95\% (95/100) of the forward steps taken in the encoding. At 95\% of the forward process, the forward step has not converted the image into a full isotropic Gaussian distribution. By retaining those last few steps in the forward process, the content of the image retains its influence when combined with the text prompt. This can be seen as for most cases in \autoref{fig:skull2dog}, by not completing the full forward process, the dogs face in the direction of the skull. It is only when the full forward process is taken, that the source image loses its influence on the target image. This can be seen in the dogs' faces changing directions to face forward. While reducing the steps taken in the forward process helps maintain faithfulness to the source image in most cases, the Revive-2I method is not always perfect. At each fraction of the forward process, the diffusion process will sometimes result in the full body of the dog and as the amount of steps taken in the forward process decrease, the faithfulness to the target domain also decreases. This can be seen in some images resulting in dog images that look like stickers, puppets, or masks. Additional failure cases can be found in \autoref{append:fail}.

\section{Prompt Analysis}
The initial prompt, ``a photo of the head of <class\_name>," was used to mimic the class level constraint used in DDIB~\cite{su2023dual}, but slightly constrain the generation process to animal heads only. We explore the following additional prompts with the same set up on the Revive-2I$_{95\%}$ model: \textbf{1)} "A photo of the head of a dog", a generalized version of the class name prompt (Generic), \textbf{2)} "<Class\_name>", to provide only the class-level guidance provided by DDIB (Class), and \textbf{3)} "<Class\_name> head", to remove the context of "a photo" (+head). 

\begin{table}[t]
    \centering
    \caption{Results of different prompting techniques on visual quality and classification.}
    \begin{tabular}{c|c|c|c|c}
         & FID $\downarrow$ & KID$\downarrow$ & All @1$\uparrow$ & Class @1$\uparrow$\\
         \hline
        DDIB~\cite{su2023dual}  & 204.42 & 0.0698 & 73.68 & 56.14\\
        Revive-2I$_{95\%}$ & \textbf{143.29} & 0.0345 &  \textbf{100} & \textbf{92.56} \\
        \hline
        Generic & 183.16 & 0.0965 & 88.43 & 1.65 \\
        Class & 155.83 & 0.0391 & 89.69 & 80.41 \\ 
        +head & 160.50 & \textbf{0.0334} & 80.16 & 64.46
    \end{tabular}
    \label{tab:prompting}
\end{table}

\begin{figure}
    \centering
    \caption{Results of different prompts. \textbf{Rows}:(top to bottom) Boston terrier, Boxer, Chihuahua, Great Dane, Pekingese, Rottweiler}
    \includegraphics[height=.4\textheight]{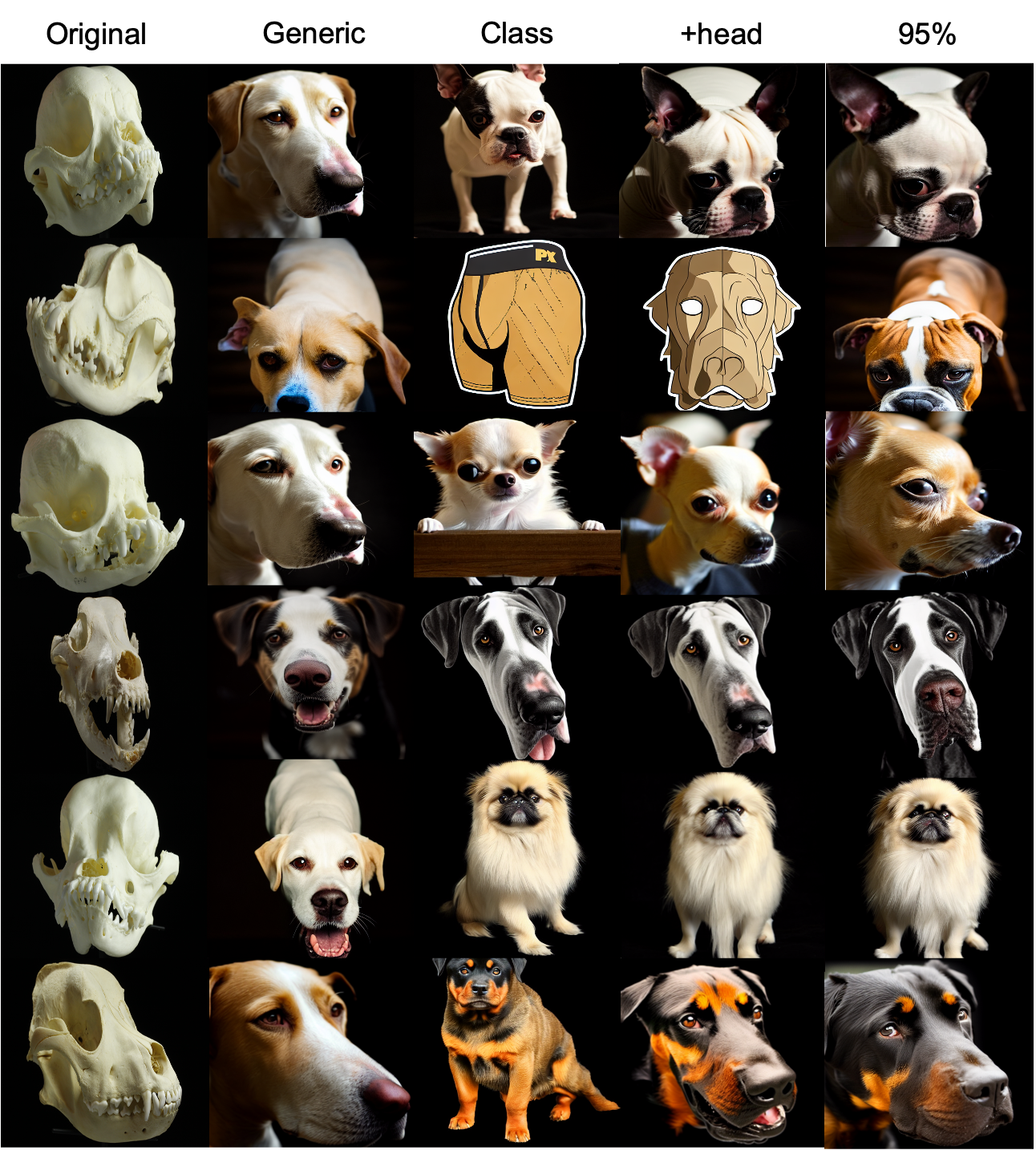}
    \label{fig:prompting}
\end{figure}

As seen in \autoref{fig:prompting}, the prompts constrain the generation process in different ways. The generic prompt generates images of the most popular dogs with the most classifications being Labrador retriever (39 images). While this prompt is a good generalization and shows that Revive-2I can be used in abstract cases, it also demonstrates that the more popular a domain is the more it might dominate a result and leave out underrepresented, and possibly correct, target images. The class name prompt provides more freedom in the generation process producing better image quality scores than the other ablation prompts. However, the freedom given in the prompt does not always constrian the generation process to dogs (the underwear boxers generated for a boxer) or allows for the generation of additional artifacts (the wooden board visible in chihuahua). While the +head prompt does not constrain the generation process to living dogs, producing images commonly classified as masks or stickers.




\section{Limitations}
One limitation is faithfulness to the source image. Deviations from the source image can be seen when the animal faces a different direction than the skull or when additional artifacts like the animals body are generated. This unfaithfulness comes from the strong understanding of the target domain provided by guidance. While this helps the model reach the target domain, it sometimes provides too much information. It is challenging to provide the correct amount of information to not generate additional features or not enough information to fully translate the image. 

Another limitation is in producing unseen classes. For example, when translating the bones of dinosaurs it would be helpful to guide the translation process with "a photo of a mammal". This was attempted with "a photo of a dog head", however, this only produced the most popular breeds like Labrador retrievers. This result is undesirable in cases where the translation class is unknown and needs generalization.

\section{Conclusion}
In this paper, we propose the task of translating skulls into living animals (Skull2Animal). The task requires the generation of a large number of new visual features, inference about geometry in the target image, and provides a verifiable constraint on the translation process. We show that traditional I2I methods using GANs are not able to successfully bridge the domain gap and instead propose the use of guided diffusion models. By providing a classifier or text prompt, we can encode more information about the target domain into the diffusion process. However, our method still lacks full faithfulness to the source image. Thus, being able to encode enough information about the target domain but retaining the intrinsic source context, like head direction, is an interesting direction for future research. 

\section{Acknowledgements}
We greatly appreciate Nick Mann for allowing for the use of his skull photos. Thanks Kurtis for his discussion. This work was supported in part by the Georgen Institute for Data Science at the University of Rochester.

\clearpage


\bibliographystyle{ACM-Reference-Format}
\balance
\bibliography{sample-base}

\clearpage


\appendix
\section{Dataset Statistics}
\label{append:data}
\begin{table}[t]
    \centering
    \caption{Breakdown of animal distributions in Skull2Animal.}
    \begin{tabular}{c|c|c}
        Wild Animal & Living Images & Skull Images\\
        \hline
        Kit fox & 84 & 152\\
        Fennec fox & 0 & 201\\ 
        Red fox & 281 & 201\\
        Artic fox & 55 & 0\\
        Grey fox & 44 & 0\\
        \hline
        \textbf{Fox} & \textbf{464} & \textbf{554} \\
        \hline
        Clouded leopard & 0 & 291\\
        Leopard & 477 & 192 \\
        Snow leopard & 282 & 0 \\
        Jaguar & 253 & 0\\
        Cheetah & 665 & 242\\
        \hline
        \textbf{Leopard} & \textbf{1678} & \textbf{726}\\
        \hline 
        Boston terrier & 22 & 222\\
        Boxer & 71 & 198\\ 
        Chihuahua & 215 & 168\\
        Great dane & 27 & 204\\
        Pekinese & 26 & 199\\
        Rottweiler & 61 & 210\\
        Other (random) & 779 & 0\\
        \hline
        \textbf{Dogs} & \textbf{1201} & \textbf{1201}\\
        \hline
        \textbf{Wild Animals} & \textbf{3343} & \textbf{2481}
    \end{tabular}
    \label{tab:wild_count}
\end{table}
The Skull2Animal dataset can be broken down into 4 species: cats, dogs, foxes, and leopards (\autoref{tab:wild_count}). Because this dataset was built off of AFHQ \cite{choi2020stargan}, it only has mammals. The skull portion of the dataset contains many more types of animal species. To request additional skull images or for inquiries about the data, please reach out to the first author or visit the project page \url{https://tinyurl.com/skull2animal}.



\section{Parameter Settings}
To understand more about the performance of the method, we explore different levels of classifier free guidance (cfg). We find similar results to Rombach et al~\cite{rombach2022highresolution}, that increasing the cfg increases the alignment between image and text, but increasing the guidance beyond a certain threshold yields a decrease in image quality. This can be seen in \autoref{tab:parameters}, as the original Revive-2I model (cfg scale of 7.5), yields the best alignment with the class labels for each skull while also having the lowest FID. 
\begin{table}[h]
    \centering
    \begin{tabular}{c|c|c|c|c}
         & FID $\downarrow$ & KID$\downarrow$ & All @1$\uparrow$ & Class @1$\uparrow$\\
         \hline
        Revive-2I$_{95\%}$ & 143.29 & 0.0345 &  \textbf{100} & \textbf{92.56} \\
        \hline
        Scale 5.0 & 146.78 & 0.0345 & \textbf{100} & 84.30 \\
        Scale 6.0 & 143.93 & 0.0340 & \textbf{100} & 88.43 \\
        Scale 7.0 & 148.05 & 0.0341 & \textbf{100} & 90.91 \\
        Scale 8.0 & 148.40 & 0.0352 & \textbf{100} & 90.91 \\
        Scale 9.0 & 152.14 & 0.0340  & \textbf{100} & 90.91\\
        Scale 10.0 & 151.31 & 0.0370 & \textbf{100} & 90.91
    \end{tabular}
    \caption{Different classifier free guidance scales.}
    \label{tab:parameters}
\end{table}

\section{Small Domain Gap Translations (shortI2I)}
\label{append:shortI2I}
To demonstrate the ability of the Revive-2I method, we extend the study to the shortI2I, image-to-image translation across small domain gaps, and the task of translating pictures into paintings by Claude Monet using the Monet2Photo dataset\footnote{The dataset was obtained from \url{https://www.kaggle.com/datasets/balraj98/monet2photo}} from CycleGAN \cite{zhu2020unpaired}. As previously done in \autoref{sec:method}, we use a variety of forward process lengths. For this ablation, we use $T=100$ forward steps and use $\frac{T}{4}$ (25), $\frac{T}{2}$ (50), $\frac{3T}{4}$ (75), and $T$ (100) forward steps. Each method is guided by the same prompt `a painting by Claude Monet' on the same images. 

\begin{figure*}
    \centering
    \includegraphics[width=\textwidth, height=.25\textheight]{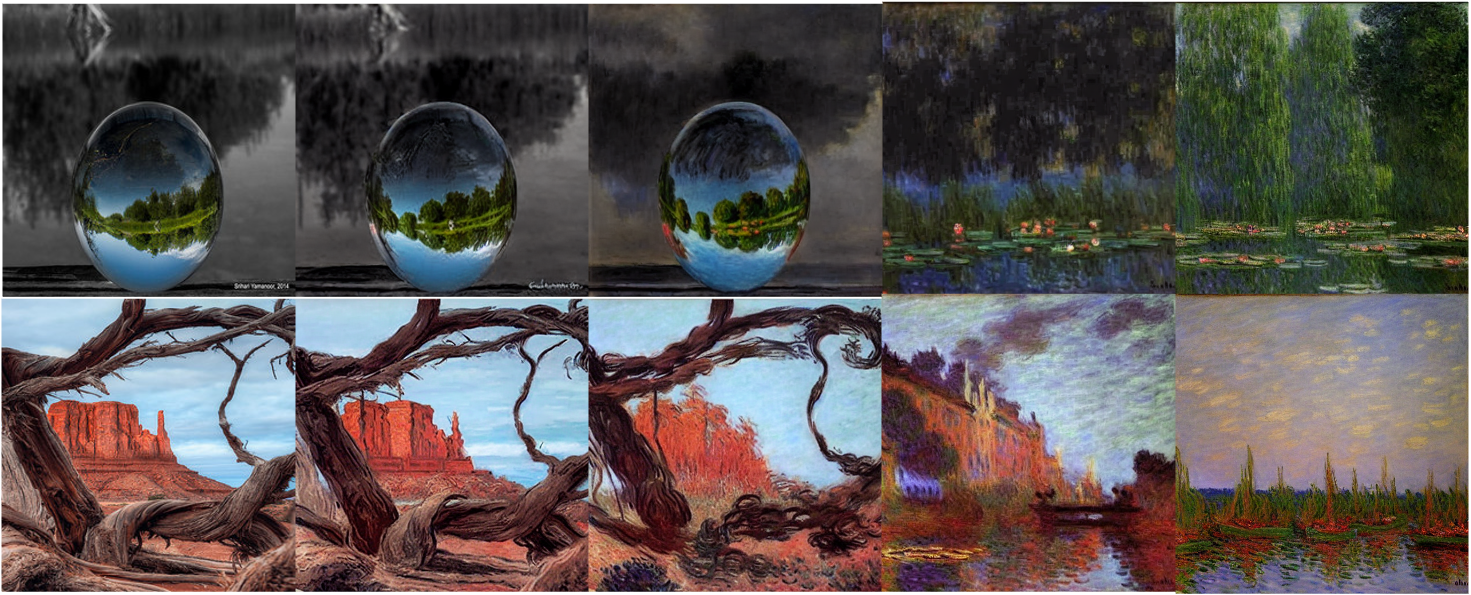}
    \caption{Revive-2I on Monet2Photo (photo$\to$Monet): \textbf{Columns:} Input, 25\%, 50\%, 75\%, 100\%.}
    \label{fig:monet2photo}
\end{figure*}

As can be seen in \autoref{fig:monet2photo}, Revive-2I is capable of doing shortI2I, however some steps perform better than others. It would appear that $\frac{T}{4}$ is too few steps, preserving the content of the image, but not transferring the style of the Monet photos. Steps $\frac{3T}{4}$ and $T$ are too many in the forward process, losing the intrinsic source features and instead generating new photos. With $\frac{T}{2}$ forward steps, the image content is well preserved and style is transferred into the Monet painting domain, demonstrating the methods ability to provide global edits to images like null-text inversion \cite{mokady2022nulltext}.

\section{Failure Cases}
\label{append:fail}
In this section we evaluate the failure cases of Revive-2I's diffusion step combinations for 50\% and 95\%. We only evaluate these two cases because we find that 50\% to 70\% are very similar in results and 95\% was the best version of the model found is \autoref{sec:results}. 

\begin{figure*}
    \centering
    \includegraphics[width=\textwidth, height=.12\textheight]{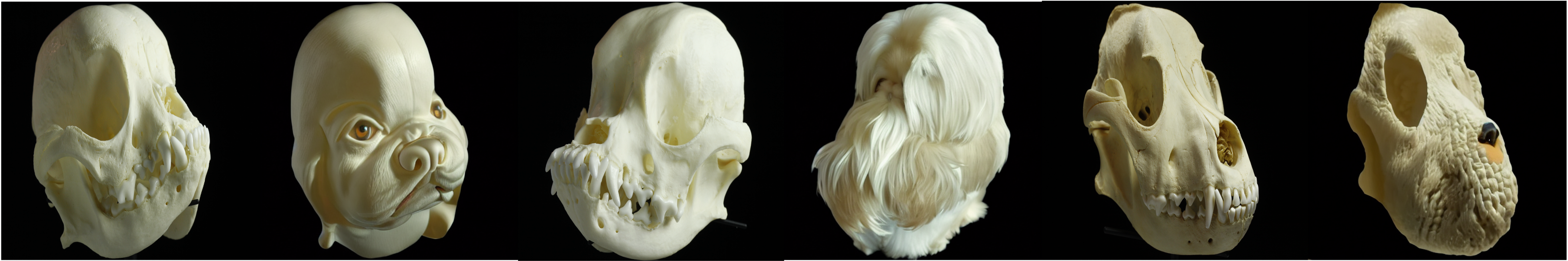}
    \caption{Failures of Revive-2I$_{50\%}$ on Skull2Dog. In order: Boston Terrier, Boxer, Rottweiler.}
    \label{fig:50_skull2dog}
\end{figure*}

\paragraph{50\%:} In terms of faithfulness to the source image, Revive-2I$_{50\%}$ retains the most of the source content. In \autoref{fig:50_skull2dog}, it can be seen that the living dogs face the same way as the skull for each breed. However, the quality of the generated dogs is poor. These results can be accounted to the source content playing to great of an influence in the translation process. This result is good in cases of shortI2I (\autoref{append:shortI2I}), but when a large amount of visual features need to be generated, more influence from the text prompt is needed. It may be important to note that the errors of Revive-2I$_{50\%}$ are very similar to the results of the image editing technique null-text inversion~\cite{mokady2022nulltext}.

\begin{figure*}
    \centering
    \includegraphics[width=\textwidth, height=.12\textheight]{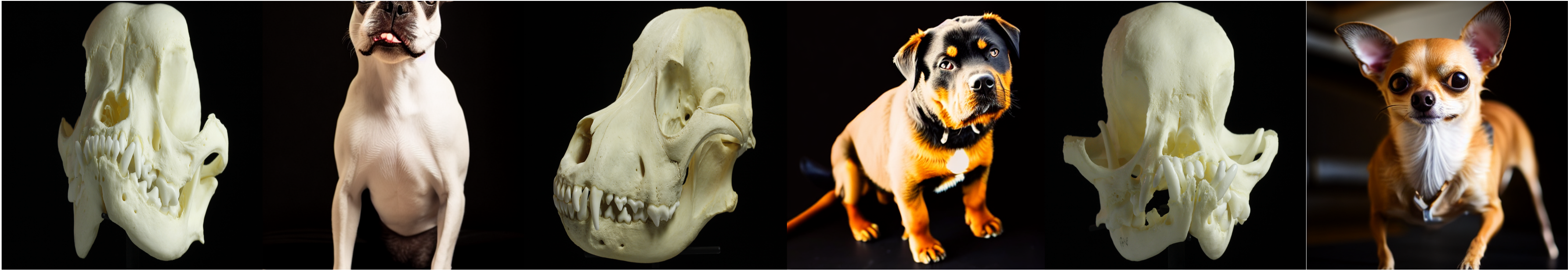}
    \caption{Failures of Revive-2I$_{95\%}$ on Skull2Dog. In order: Boston Terrier, Rottweiler, Chihuahua.}
    \label{fig:95_skull2dog}
\end{figure*}

\paragraph{95\%:} While Revive-2I$_{95\%}$ is able to create the visual features needed, it sometimes encounters the same problems as DDIB's \cite{su2023dual} classifier guidance where the full body image of the dog is created (\autoref{fig:95_skull2dog}). This is error is most likely caused in the cross-attention between the prompt and latent in the UNet layers. We leave this exploration for future work.

\section{Leopard and Fox Datasets}
\begin{figure*}
    \centering
    \includegraphics[width=\textwidth]{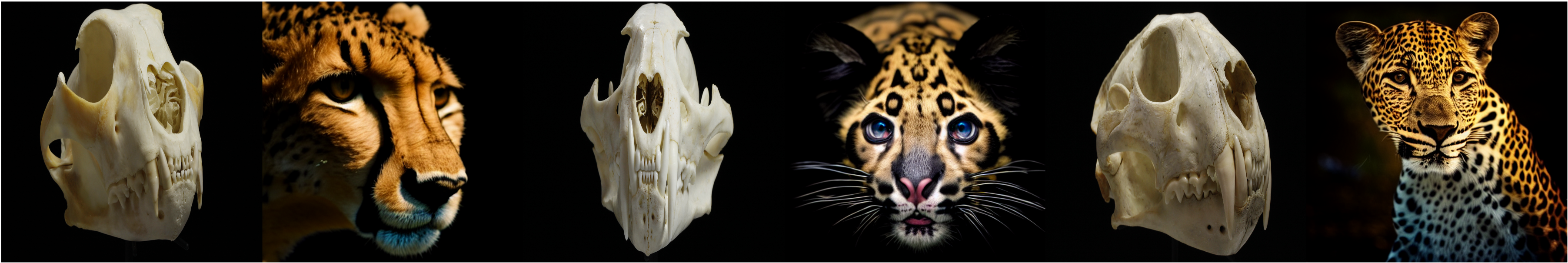}
    \caption{Revive-2I$_{95\%}$ on Skull2Leopard. In order: Cheetah, Clouded Leopard, Leopard.}
    \label{fig:95_leopard}
\end{figure*}

\begin{figure*}
    \centering
    \includegraphics[width=\textwidth]{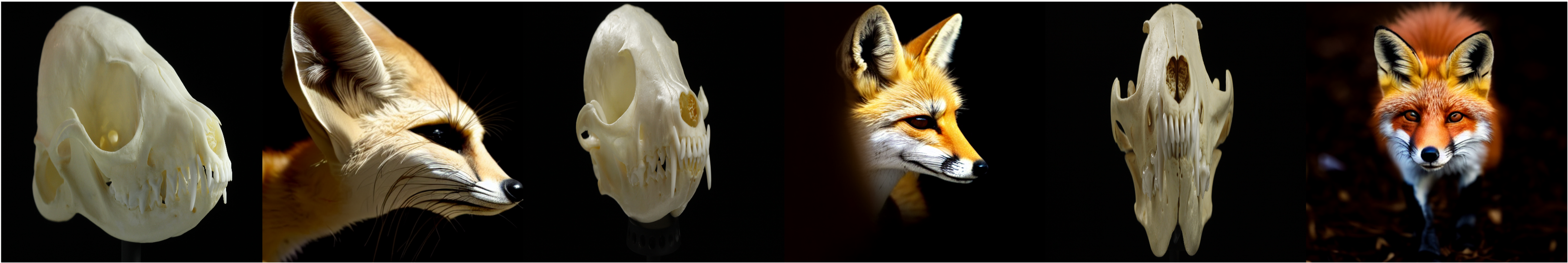}
    \caption{Revive-2I$_{95\%}$ on Skull2Fox. In order: Fennec Fox, Kit Fox, Red Fox.}
    \label{fig:95_fox}
\end{figure*}

Lastly, we show the Revive-2I$_{95\%}$ model on the leopard and fox datasets in \autoref{fig:95_leopard} and \autoref{fig:95_fox}. The prompt used for the fox and leopard generations is ``a photo of the head of" and the ImageNet class (the same prompt as \autoref{sec:method}). For example, the clouded leopard prompt is ``a photo of the head of a clouded leopard" and the fennec fox is ``a photo of the head of a fennec fox".

\end{document}